\documentclass[letterpaper, 10 pt, conference]{ieeeconf}  

\IEEEoverridecommandlockouts                              

\usepackage{multirow}
\usepackage{xurl}
\usepackage{hyperref}

\usepackage[style=ieee, citestyle=numeric-comp, backend=biber]{biblatex}
\usepackage{stfloats}
\usepackage{booktabs}
\usepackage{tabularx}
\usepackage{threeparttable}
\newcommand{\citet}{\cite}

\usepackage{caption}
\usepackage{graphicx}
\usepackage{etoolbox}
\newcommand{\insertfig}{
    \includegraphics[width=\linewidth]{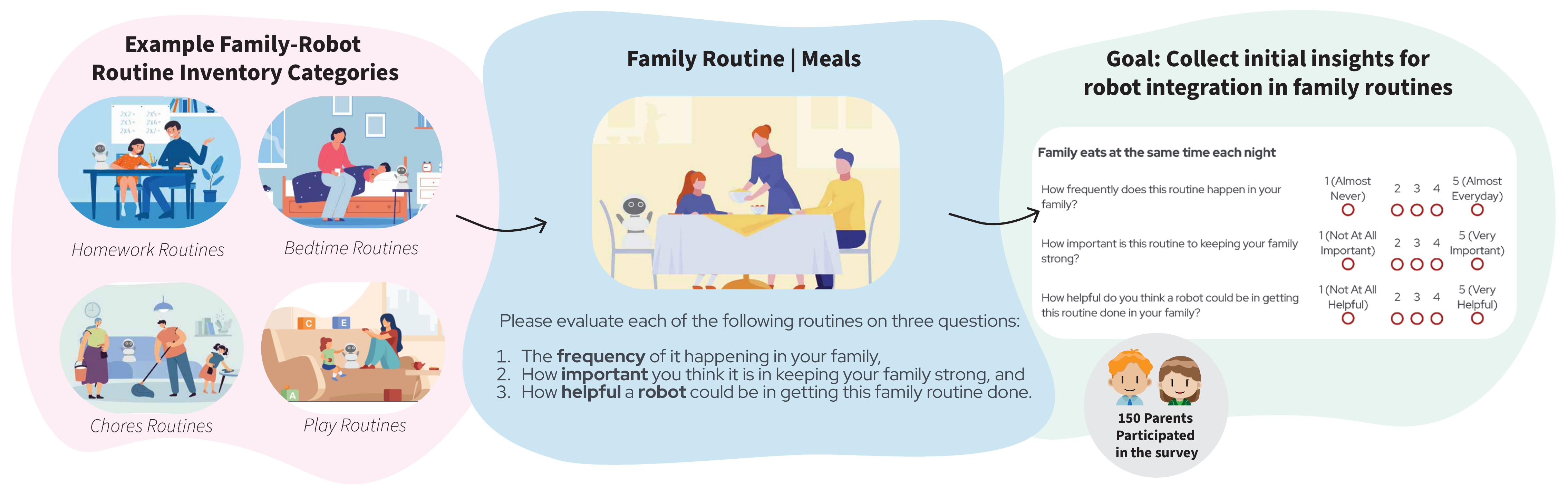}
    \captionof*{figure}{
        \phantomsection
        \label{fig:teaser}
        Fig. 1: 
        \textbf{Family-Robot Routines Inventory:} A survey to collect initial insights for robot integration in family routines.
    }
}

\makeatletter
\apptocmd{\@maketitle}{\centering\insertfig}{}{}
\makeatother

\title{\LARGE \bf
Robots in Family Routines: Development of and Initial Insights from the Family-Robot Routines Inventory
}

\author{Michael F. Xu,$^{1}$ Bengisu Cagiltay,$^{1}$  Joseph Michaelis,$^{2}$  Sarah Sebo,$^{3}$ Bilge Mutlu$^{1}$
\thanks{$^{1}$Department of Computer Sciences, University of Wisconsin--Madison, Madison, WI, USA; email:
        {\tt\small \{michaelfxu, bengisu, bilge\}@cs.wisc.edu}}
\thanks{$^{2}$Department of Computer Science, University of Illinois Chicago, Chicago, IL, USA; email:
        {\tt\small jmich@uic.edu}}%
\thanks{$^{3}$Department of Computer Science, University of Chicago, Chicago, IL,
USA; email:
        {\tt\small sarahsebo@uchicago.edu}}%
}

\begin{document}

\maketitle
\thispagestyle{empty}
\pagestyle{empty}

\begin{abstract}

Despite advances in areas such as the personalization of robots, sustaining adoption of robots for long-term use in families remains a challenge. Recent studies have identified integrating robots into families' routines and rituals as a promising approach to support long-term adoption. 
However, few studies explored the integration of robots into family routines and there is a gap in systematic measures to capture family preferences for robot integration.
Building upon existing routine inventories, we developed \textit{Family-Robot Routines Inventory (FRRI)}, with 24 family routines and 24 child routine items, to capture parents' attitudes toward and expectations from the integration of robotic technology into their family routines. Using this inventory, we collected data from 150 parents through an online survey. 
Our analysis indicates that parents had varying perceptions for the utility of integrating robots into their routines. For example, parents found robot integration to be more helpful in children's individual routines, than to the collective routines of their families.
We discuss the design implications of these preliminary findings, and how they may serve as a first step toward understanding the diverse challenges and demands of designing and integrating household robots for families.

\end{abstract}

\section{INTRODUCTION}

In recent years, household robots have been increasingly introduced to both the research community and the consumer market. Robots, as physically embodied social agents, can create value to their users in various scenarios. Working with individual family members, robots can help in areas such as personalized tutoring to children~\cite{belpaeme2018social,ramachandran2019personalized}, and assistance and companionship to the older adults~\cite{gross2015robot,lee2018reframing}. At the same time, robots have also demonstrated potential in improving interactions among family members~\cite{chen2022designing, ho2024s}, including interactions among inter-generational members of the family~\cite{short2017understanding, joshi2019robots}.
While these applications have sparked an increased interest in the integration of robots in family lives, many challenges remain. For example, users often report disappointment in the robot's capabilities \cite{fernaeus2010you,ostrowski2022mixed}, and find the robots lacking of practical benefits and usefulness \cite{caudwell2019ir}. As a result, interactions between the human adopter and the robot generally starts decreasing after a few weeks, and often completely ceasing before the end of six months \cite{ostrowski2022mixed, de2017they,weiss2021merely}. These challenges in household-robot integration hinder the ability to fully realize the value proposition delivered by social robots.
Researchers have attempted to address this challenge in various ways, such as creating more dynamic capabilities and content~\cite{leite2013social}, improving customization and personalization~\cite{lee2012personalization,irfan2019personalization}, and developing and endowing robots with more sense of personalities~\cite{weiss2021merely}. However, sustaining adoption and interaction over the long-term still remains a challenge, and there is limited guidance in HRI that can inform ways for robot integration into family life.

We argue that, a possible solution to supporting long-term human-robot interaction could be through integrating social robots into current or future \textit{family routines and rituals}~\cite{cagiltay2024supporting}. For example, a recent study on owners and their family robots has revealed that there could be a positive feedback cycle between habitual everyday interactions\textit{---i.e., rituals---}and integration of robots into the owners' daily routines~\cite{kamino2024constructing}:

\begin{quote}
\textit{By establishing daily interaction rituals with the robots, owners incorporate the robots into their existing meaningful moments, making them an integral part of their routines. They also create additional special moments with the robots, often inviting other social actors (e.g., family members), that provide additional moments of connection that would not exist without the robots.}
\end{quote}

These findings are consistent with previous works that identified the ease of integrating robots into families' routines as a key factor that influences robot acceptance~\cite{cagiltay2020investigating, cagiltay2022exploring, de2019would}.

However, there is little existing literature explicitly studying the integration of robotic technologies into family routines, as well as limited evaluation \textit{methods and metrics} that can guide design decisions for robot integration into family routines. For example, what routines are important to families? 
What routines would family members desire to integrate a robot in? Which family routines should robot designers prioritize building around? 
Should designers aim to integrate robots into current family routines, or introduce new routines enabled by the robots? What is the trade-off between family desirability and technical feasibility in designing for robot-facilitated family routines? 
As a first step toward addressing these questions, we've built upon existing routines inventories \cite{jensen1983family, sytsma2001development, grinnell2023young, fiese1993development}, and developed a survey to capture parents’ initial attitudes on integrating robots into family routines. We refer to this as the \textbf{Family-Robot Routines Inventory (FRRI)}. We conducted an online survey with over 150 parents, located across the United States, who are the caregiver of children aged between 5 and 16. Our study serves as a first step in understanding the early challenges and opportunities with integrating robots into families' routines, where FRRI can be used as a resource to systematically discover user adoption perceptions, and identify robot-facilitated routines of high priority.

\section{Background}

Below, we review the literature on (1) existing family routines inventories,  and (2) family routines and social robots.

\subsection{Existing Family Routines Inventories}

There are several measurement and metrics developed to assess the state of routines in family life, including the Family Routines Inventory (FRI) \cite{jensen1983family}, Child Routines Inventory (CRI) \cite{sytsma2001development}, Young Adult Routines Inventory (YARI) \cite{grinnell2023young}, and Family Rituals Questionnaire (FRQ) \cite{fiese1993development}.
FRI consists of 28 items, grouped into 10 categories, such as work day routines, meals, and disciplinary routines. FRI demonstrated decent reliability, and have been widely used as an assessment of family routines. However, it also has many limitations, such as some routine items being outdated, and that it primarily focuses on the whole family while leaving out routines involving individual family members.
CRI addresses some of these limitations, and consists of 36 items intended for parents to report on their children's routines. YARI was developed to serve as a measure of routines for young adults to self-report on, and consists of over 20 items under four sub-scales: daily routines, social routines, time management, and procrastination. Finally, FRQ is a 56-item questionnaire that provides a means to assess family rituals across seven settings, including dinner time, vacations, weekends, etc.
In Section \ref{section:method}, we will elaborate on our process for adapting these existing inventories and questionnaires, for developing our proposed Family-Robot Routines Inventory. 

\subsection{Family Routines and Social Robots}

For the purpose of this study, we will consider robots that are generally considered \textit{social robots}, that prioritize human interaction. Family-robot interactions, and specifically in the context of family routines, are understudied. However, many studies touch on common family routine items, including ones covered by our Family-Robot Routines Inventory. Berrezueta et al. studied robotic assistant and assessed its effectiveness in supporting children's homework activities~\cite{berrezueta2021assessment}. Through a triadic story-telling activity, Chen et al. showed that social robots can enhance parent-child interaction by engaging parents during story times~\cite{chen2022designing}. Similarly, in family shared recreational activities, Kim et al. showed that when family members play together with robots, verbal and physical activities increase compared to dyadic child-robot interactions~\cite{kim2022can}.
Previous research has also explored families' collaborative learning with social robots~\cite{ahtinen2023robocamp} and the potential role of robots' involvement in a family's bedtime routine, in the context of taking care of the robot~\cite{cagiltay2022exploring}.
As seen from these applications, social robots play many different roles in family-robot interactions. Children and parents may have varying expectations of in-home social robots~\cite{cagiltay2020investigating}. These conflicting needs pose a challenge for long-term in-home robot adoption~\cite{de2016long, de2015living, Cagiltay_engagement_2022}. Thus, a family-centric, holistic understanding of family needs and perceptions toward robots is necessary~\cite{cagiltay2024toward}.

Limited, but growing work in human-robot interaction focuses on approaches to design social robots for families. In a co-design workshop involving children and their parents, Zhang et al. gathered design preferences for a social robot in the context of pain management, which started with an introduction of the challenge and demos of several social robots~\cite{zhang2022understanding}. In another co-design activity, Thiessen et al. involved children and their family in creating their own robot prototypes, which served as a tool that facilitated discussions about challenges participants expected in integrating social robots into their daily lives~\cite{thiessen2023understanding}. Ostrowski et al.  explored co-designing with and for the older population~\cite{ostrowski2021long}. 
While each study focuses on integrating social robots into a different aspect of family activities, a common theme for robot integration is the need for a shared reference on which to base the discussions and interviews between the researchers and participants. By gathering information on both a family's routines and their attitudes on involving a robot in the various routines, our proposed Family-Robot Routines Inventory serves as a tool to base further investigations on. 

\section{Method}
\label{section:method}
We illustrate in Figure~\hyperref[fig:teaser]{1} our approach at a high-level. Below we describe (1) how we developed the survey, (2) the participants and data collection method, and (3) our data analysis methods. The full inventory of routine items, example survey screenshots, and response data on these routine items can be accessed through the repository.\footnote{\url{https://osf.io/7zha5/}}

\subsection{Survey Development}
Our developed survey consists of introducing and defining the concept of routines and social robots, a training module, routine inventory items, and open-ended questions. 

\subsubsection{Routines Definition}

We categorized the main routine items in two sections: Family Routines and Child Routines. Following existing literature, we provided the participants with definitions for \textit{Routines} and \textit{Family Routines}: ``\textit{Routines are events that occur at about the same time, in the same order, or in the same way every time}'' \cite{jordan2003further}, whereas family routines are ``\textit{observable, repetitive behaviors which involve two or more family members and which occur with predictable regularity in the day-to-day and week-to-week life of the family}'' \cite{jensen1983family}. We repeated these definitions throughout the survey as reminders to the participants.

\subsubsection{Social Robot Definition}
\begin{figure}[t]
  \vspace{6pt}
  \centering
  \includegraphics[width=\linewidth]{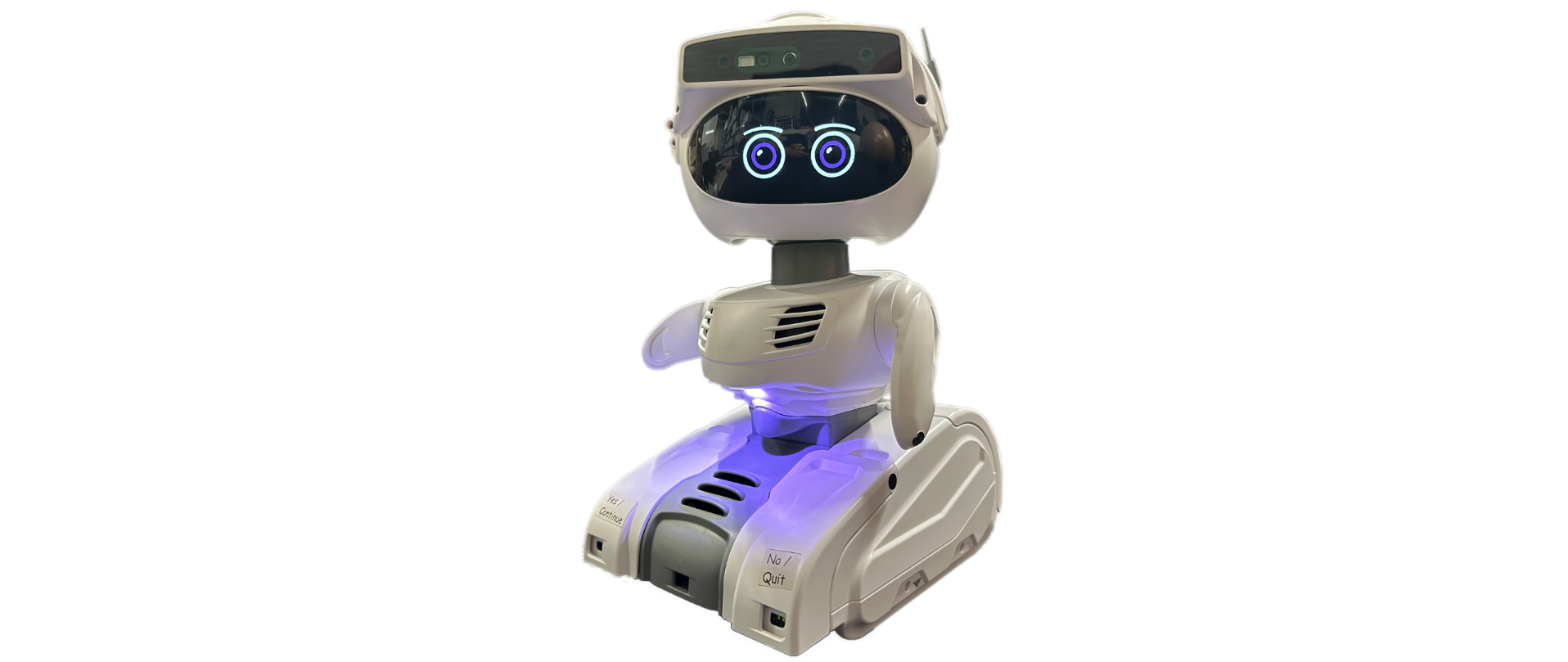}
  \caption{
    Misty II by Misty Robotics, the example robot shown to participants.
}
  \label{fig:example_robot}
  \vspace{-6pt}
\end{figure}

As there are many categories of robots with various capabilities, it is important to help the participants form an appropriate mental model of the specific type of robot we are interested in learning about. To that end, we provided an example robot figure, similar to the one shown in Figure~\ref{fig:example_robot}, along with a few lines of descriptions about its size, appearance, capabilities and limitations. Specifically, our description referred to a typical social robot, whose strength is in interacting with family members and children via verbal and non-verbal expressions, as opposed to a traditional vacuum robot, or robots that are designed to be more manipulative of physical items. An excerpt from our description reads: \textit{It is roughly the size of a cat, and it can move around the house on its own. It can listen, and speak, and overtime, it can learn to recognize the members in your household. While it has arms that can swing and rotate, it could be challenging for Misty to grab, hold, or move items very well.}

\subsubsection{Training Module}
To familiarize the participants with the survey questions, we included a training module and presented an example routine item with the questions in the identical format as they will appear in the main module.

\subsubsection{Routines Questionnaire}
\label{subsubsection:main_module}
We combined items from existing inventories focusing on routines for families \cite{jensen1983family, fiese1993development}, children \cite{sytsma2001development}, and young adults \cite{grinnell2023young}.
To update the past inventories, we made modifications to the wording, structure, and presentation of the survey. This exercise resulted in 48 routine items, with the majority originating from FRI~\cite{jensen1983family} and CRI~\cite{sytsma2001development}. These were then separated into two categories: family routines (involving two or more members), and child routines (involving an individual child). Routines are further broken up into 11 subcategories (seven in family routines, and four in children routines). To reduce fatigue for the participants, we consolidated categories in the survey. We presented participants with a total of nine categories, five in family routines, and four in child routines. To present these routine items and help participants consider the questions, we provided an illustration of a potential family-robot integration scenario for each of the nine categories. For each of the routine items, we asked the three following questions:

\begin{enumerate}
  \item How \textbf{frequently} does this routine happen in your family?
  \item How \textbf{important} is this routine to keeping your family strong?
  \item How \textbf{helpful} do you think a robot could be in getting this routine done in your family?
\end{enumerate}

We will refer to these three questions as the Frequency, Importance, and Helpfulness questions. The first two questions are directly taken from the Family Routines Inventory \cite{jensen1983family}, whereas we've constructed the third question to capture the participant’s attitudes toward integrating robotic technology into the specific routine item \textit{in their family}. For all three question types, we collected answers on 5-point anchored Likert scales. For the Frequency questions, options ranged from ``1 (Almost Never)'' to  ``5 (Almost Everyday).'' Options for the Importance questions ranged from ``1 (Not At All Important)'' to ``5 (Very Important).'' Finally, the Helpfulness scale ranged from ``1 (Not At All Helpful)'' to ``5 (Very Helpful).''

After completing the routine questionnaire, participants received open-ended questions to list any routines in their family that have not been covered by the current inventory, as well as any comments or concerns.
Finally, participants filled a demographic questionnaire to list basic demographic information of themselves and family members.

\begin{table}[t]
  \vspace{2pt}
  \caption{Demographic Information of Participants}
  \label{tab:participants}
  \begin{center}
  \begin{threeparttable}

  \begin{tabularx}{\linewidth}{Xcc}
    \toprule
     &Frequency&Percentage$^*$\\
    \midrule
    \textbf{Role in Family} & & \\
    \hspace{1mm} Mother & 85 & 56.7 \\ 
    \hspace{1mm} Father & 65 & 43.3 \\ 

    \textbf{Age} & & \\
    \hspace{1mm} min & 23 &  \\ 
    \hspace{1mm} max & 61 &  \\ 
    \hspace{1mm} median & 40.0 &  \\


    \textbf{Race} & & \\
    \hspace{1mm} White & 99 & 66.0 \\ 
    \hspace{1mm} Black or African American & 31 & 20.7 \\ 
    \hspace{1mm} Asian & 9 & 6.0 \\ 

    
    \hspace{1mm} Other & 5 & 3.5 \\ 


    \textbf{Employment} & & \\
    \hspace{1mm} Full-time & 101 & 67.3 \\ 
    \hspace{1mm} Part-time & 14 & 9.3 \\ 
    \hspace{1mm} Not Working & 22 & 14.7 \\ 

    
    \textbf{Primary Caregiver in Family} & & \\
    \hspace{1mm} The Participant & 75 & 50.0 \\ 
    \hspace{1mm} The Other Parent & 13 & 8.7 \\ 
    \hspace{1mm} Equally & 62 & 41.3 \\ 


    \textbf{Number of Children} & & \\
    \hspace{1mm} 1 & 69 & 46.0 \\ 
    \hspace{1mm} 2 & 60 & 40.0 \\ 
    \hspace{1mm} 3 & 18 & 12.0 \\ 
    \hspace{1mm} 4 & 3 & 2.0 \\

    \textbf{Total Number of Children} & 255 & \\
    
    \textbf{Children Age Range} & & \\
    \hspace{1mm} 1 - 4 & 7 & 2.7 \\ 
    \hspace{1mm} 5 - 8 & 79 & 31.0 \\ 
    \hspace{1mm} 9 - 12 & 94 & 36.9 \\ 
    \hspace{1mm} 13 - 16 & 75 & 29.4 \\ 

    \textbf{Children Gender} & & \\
    \hspace{1mm} Male & 131 & 51.4 \\ 
    \hspace{1mm} Female & 124 & 48.6 \\ 

  \bottomrule
  
\end{tabularx}
  
\begin{tablenotes}
    \item $^*$Percentages may not add up to 100, as some participants did not provide relevant responses to all questions.
\end{tablenotes}

\vspace{-12pt}
\end{threeparttable}
\end{center}
\end{table}

\subsection{Participants and Data Collection}

We recruited 161 participants through the online research platform Prolific\footnote{\url{https://www.prolific.com}} and hosted the survey on Qualtrics\footnote{\url{https://www.qualtrics.com}}. The inclusion criteria were: participants that are 18 years or older, residing within the United States, and fluent in English. Furthermore, participants were included if they were a primary caregiver of at least one child between the age of 5 to 16 who lives within the same household as the participant at least 5 days a week.
The survey estimated to take 25--30 minutes and the median duration among all participants were about 21 minutes. Participants received \$6 USD for completing the survey. Five pilot participants and six participants that failed the attention check were excluded from the data. The final set included data from 150 participants. 
Table \ref{tab:participants} reports the demographic information of the 150 participants and the children in the participants' families.

\subsection{Data Analysis}
We performed both qualitative and quantitative analysis on the survey responses. We generated descriptive statistics and correlation coefficients for the set of questions on the 48 routine items, grouped and analyzed at various levels. The first two authors conducted a Thematic Analysis~\cite{braun2006using} on responses to the open question asking about additional routines in participant families that were not covered by the survey. We first generated potential codes, and then individually assigned codes to the reported routine items. We then discussed and iterated on the code assignment until an agreement was reached for each routine item.

\section{Results}

In this section, we analyze survey responses to FRRI, and investigate patterns among routine categories, individual routine items, as well as their relationship with the three main questions we ask about each routine item: their Frequency, Importance, and Helpfulness. We also conducted a Thematic Analysis on the open question about additional routines not covered in FRRI. We present these insights in four categories. 

\begin{figure*}[t]
  \centering
  \includegraphics[width=\linewidth]{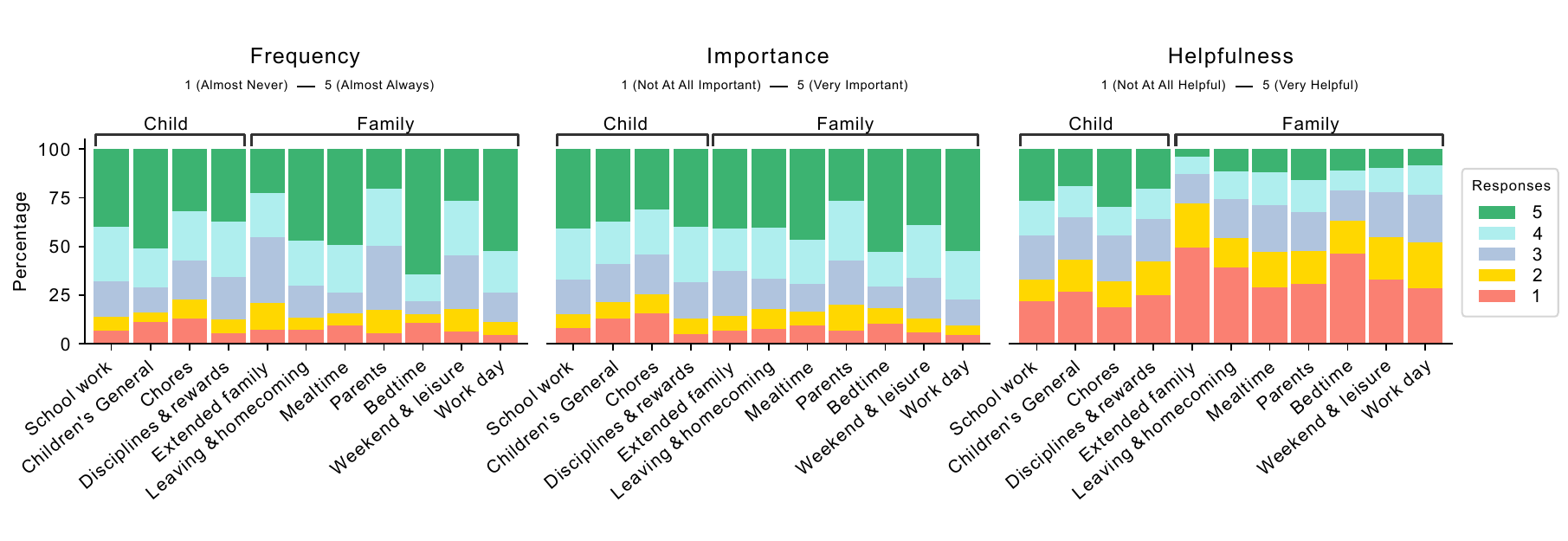}
  \caption{
    Distribution of the responses, by question types, and aggregated by routine categories.
}
  \label{fig:result_dist_by_routines}
\end{figure*}

\begin{figure}[!b]
  \centering
  \vspace{-8pt}
  \includegraphics[width=\linewidth]{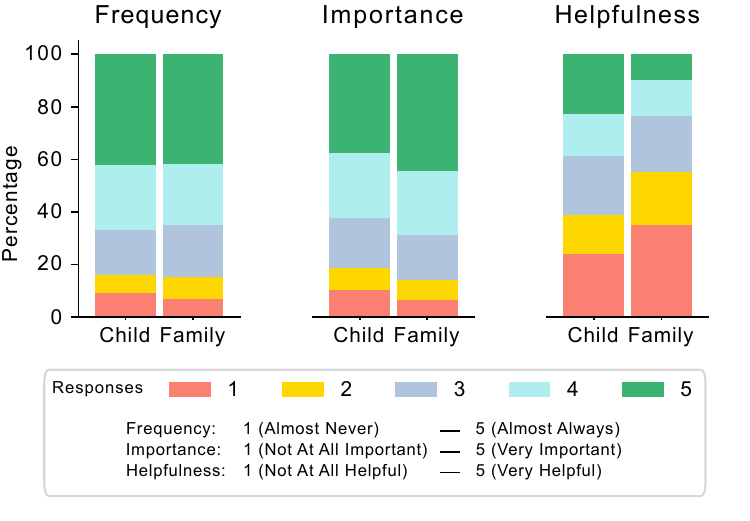}
  \caption{
    Distribution of the responses, by question types, and aggregated by Family Routines vs. Child Routines.
}
  \label{fig:result_dist_by_fam_child}
\end{figure}

\subsection{\textbf{Insight 1.} Parents found robot integration to child-routines more helpful, compared to integration into family routines.} Compared to family routines, child routine categories consistently received higher scores on parent's perceived helpfulness of involving a robot. The distribution of the participants' responses for the three question types on the various routine categories are presented in Figure~\ref{fig:result_dist_by_routines}. In Figure~\ref{fig:result_dist_by_fam_child}, responses were aggregated by family routines and child routines, each consisting of 24 routine items.

We observe that the child routines category that received the lowest Helpfulness score (\textit{Children's General Routines}, $mean = 2.84$) still received a higher score than the highest ranking family routine category (\textit{Parent's Routines}, $mean = 2.69$). This distinction is not seen in the Frequency or the Importance questions, where the scores for both Frequency and Importance mostly range from $3.5$ to $4$, for both child routines and family routines.

In Table \ref{tab:top_bottom_helpful}, we list the top and bottom five routines in terms of robot Helpfulness scores. The routine that participants found most helpful to involve a robot was ``\textit{Children studies for tests (\textit{e.g.,} weekly spelling test)},'' receiving an average of $3.47$ out of $5$, and the least being ``\textit{Family regularly visits the relatives},'' with a score of $1.7$. 
Notably, the top five routine items include only child routine categories, and four out of the bottom five capture family routines.


\begin{table}[tbp]
  \vspace{2pt}
  \caption{Routines with High and Low Helpfulness Scores}
  \label{tab:top_bottom_helpful}
  \begin{center}
  \begin{tabularx}{\linewidth}{Xc}
    \toprule
     Routine Item & Avg. Score \\
    \midrule

    \textbf{Top 5} & \\
    \hspace{1mm} 1. Children studies for tests (\textit{e.g.,} weekly spelling test.) &  3.47  \\
    \hspace{1mm} 2. Children have time limits on fun activities (\textit{e.g.,} &  3.45  \\
        \hspace{4mm} outside play, TV, video games, or phone use.) & \\
    \hspace{1mm} 3. Children must finish household responsibilities  &  3.43  \\
        \hspace{4mm} (\textit{e.g.,} homework or chores) before play time. & \\
    \hspace{1mm} 4. Children pick up toys and puts them away when &  3.43  \\
        \hspace{4mm} done playing. & \\
    \hspace{1mm} 5. Children do regular household chores (\textit{e.g.,} take &  3.41  \\
        \hspace{4mm}  out trash, helps with laundry, feeds or cares for & \\
        \hspace{4mm}   family pet.) & \\

    & \\

    \textbf{Bottom 5} & \\
    \hspace{1mm} 1. Family regularly visits the relatives.  &  1.7  \\
    \hspace{1mm} 2. Working parent(s) come home from work at the &  1.81  \\
        \hspace{4mm} same time each day. & \\
    \hspace{1mm} 3. Family goes someplace special together almost &  1.85  \\
        \hspace{4mm} every week. & \\
    \hspace{1mm} 4. Family has special things they do each night at  &  1.88  \\
        \hspace{4mm} bedtime (\textit{e.g.,} a good-night kiss.) & \\
    \hspace{1mm} 5. Young children go to play-school the same days &  1.97  \\
        \hspace{4mm} each week. & \\
    
  \bottomrule
\end{tabularx}
\vspace{-12pt}
\end{center}
\end{table}

\subsection{\textbf{Insight 2.} Parents have diverse attitudes on whether a robot could be helpful in routines.} Responses to the robot Helpfulness questions exhibit high variances across participants.
In fact, we found higher variances for Helpfulness in almost all routine categories (with values between $1.17$ and $1.48$, and overall SD of $1.44$) compared to Frequency and Importance (with values between $1.11$ and $1.42$, and overall SDs at $1.27$ and $1.28$, respectively). 
This pattern is also observable in Figure~\ref{fig:result_dist_by_routines} and Figure~\ref{fig:result_dist_by_fam_child}, where the response distributions for the Helpfulness question are visually more spread out than their counterparts in Frequency and Importance.

Note that this variance is not explained by within-participant variations, as individual participant's responses to the Helpfulness questions exhibit comparable variances. The average of participant-level SDs is $1.07$ for Helpfulness, and the corresponding participant-level averages for Frequency and Importance are even marginally higher, at $1.12$ and $1.11$, respectively. This indicates that \textit{across participants}, there is a higher level of diversity in terms of parent's attitude on whether a robot could be helpful in getting the array of routines done. 
Moreover, within the Helpfulness responses, child routines also have consistently higher variances compared to family routine categories, with SD values between $1.46$ and $1.48$, compared to values between $1.17$ and $1.45$ for family routines. The Helpfulness section in Figure~\ref{fig:result_dist_by_fam_child} demonstrates this with a more even distribution of responses for the Child routines. This reflects even more diversity among families' attitudes, when it comes to handling, and potentially involving a robot in, their children's routines.

\begin{table}[b]
  \caption{Correlation Matrix}
  \label{tab:correlation}
  \begin{center}
  \begin{threeparttable}
  \begin{tabularx}{\linewidth}{Xccc}

     \multicolumn{4}{c}{\textbf{(Participant, Routine)-Level}} \\
     \toprule
    
     & Frequency&Importance&Helpfulness\\
    \midrule
    Frequency   & 1    & & \\
    Importance  & 0.73 & 1 & \\
    Helpfulness & 0.22 & 0.21 & 1\\

    \\
    \multicolumn{4}{c}{\textbf{Participant-Level}} \\
    \toprule
    
     & Frequency&Importance&Helpfulness\\
    \midrule
    Frequency &1&&\\
    Importance &0.80&1&\\
    Helpfulness &0.22&0.31&1\\
  \bottomrule
\end{tabularx}

\begin{tablenotes}
    \item All values significant at p $<$ .001.
\end{tablenotes}
\end{threeparttable}
\end{center}
\end{table}

\subsection{\textbf{Insight 3.} The Robot Helpfulness question adds value by capturing information otherwise unaccounted for.} Responses to the Frequency and Importance questions are more highly correlated, compared to their correlations with the Helpfulness question. In Table \ref{tab:correlation}, we present both the (participant, routine)-level and the participant-level Pearson correlation coefficient matrix of the three questions. The (participant, routine)-level matrix considers each response individually, while the participant-level matrix first averages the responses to a particular question type within each of the participants. At the (participant, routine)-level, Frequency and Importance exhibit a high level of correlation at $0.73$, while their respective correlations with Helpfulness are only around $0.21$. Values are similar at the participant level. This confirms that whether or not a robot is perceived to be potentially helpful to a routine is a complex matter. While a routine being important or highly frequent may predict a higher robot Helpfulness score, a large part of its variation remains unaccounted for. At the same time, this relatively low correlation highlights the value of the Helpfulness question, as it complements the rest of the inventory with information that would not have been otherwise captured.

\subsection{\textbf{Insight 4.} There is a diverse set of additional family-specific routines that are not covered in FRRI.}

We conducted a Thematic Analysis on the responses provided to the open questions, asking participants to share additional routines in their families that have not been covered by the items in FRRI. We also asked participants to evaluate the Helpfulness and Frequency for these additional routines, just like they've done for the routines included in the main survey. We identified several common themes for additional routines. 
108 out of the 150 participants contributed at least one additional routine, with a total of 236 routine items reported. Our analysis yielded 11 main categories of routines, with \textit{Shared Recreation} accounting for about 35\% of the items reported. \textit{Social Emotional} routines at second place (13.1\%), closely followed by \textit{Care Taking}, \textit{Housekeeping}, \textit{Packing and Prepping}, and \textit{Mealtime} routines, all accounting for around 10\% of the reported items. The rest together make up the remaining 10\%. While \textit{Shared Recreation} was mentioned the most, the average Helpfulness and Frequency scores were both below average, sitting at 3.11 and 3.72, respectively. The average among all reported additional routines were 3.22 for Helpfulness, and 3.98 for Frequency. Among the six main categories described above, \textit{Housekeeping}, \textit{Schoolwork}, and \textit{Packing and Prepping} take the top three in terms of Helpfulness scores (3.92, 3.88, and 3.84), while \textit{Caretaking}, \textit{Social Emotional}, and \textit{Packing and Prepping}, take the top three for Frequency scores (4.92, 4.23, and 4.08). 

To list a few examples, some popular routines were: Taking care of pets (\textit{Caretaking}), doing exercises together (\textit{Shared Recreation}), watching movies or TV together (\textit{Shared Recreation}), preparing meals or cooking together (\textit{Packing and Prepping}, or \textit{Mealtime}), and family prayer (\textit{Social Emotional}).
These findings suggest that many families have routines that are specifically important to them. While some of them are relatively common across participants, others are more unique to the families, and would be challenging to capture even if FRRI were to be expanded.


\section{Discussion}

Building on existing routines inventories, i.e., Family Routines Inventory (FRI) and Child Routines Inventory (CRI), we developed Family-Robot Routines Inventory (FRRI) to assess families' initial attitudes on integrating robotic technology into their routines. We designed the survey in a modular fashion, so that it could be adapted to serve as an inventory to collect initial attitudes of parents towards integrating other robots or technologies into their family routines. We reported four insights based on the analysis of survey responses. Below we discuss (1) our four insights, (2) design implications, and (3) limitations and future work.

\subsection{The Four Insights on Family-Robot Integration}

Through the insights presented in the Results section, we've demonstrated the high level of variance exhibited in the responses to the robot Helpfulness question. While we've shown some patterns regarding the differential results between family routines and child routines in \textbf{Insight 1}, it was clear from the later insights that the issue becomes more nuanced once we look beyond the broad categories. In \textbf{Insight 2}, we've shown that the variance in perception of Helpfulness is high across participants, and higher than the corresponding variances in Frequency and Importance. \textbf{Insight 3} further noted that a large part of the variation in responses to the robot Helpfulness question cannot be accounted for by responses to the Frequency or Importance question. Finally, we've highlighted in \textbf{Insight 4} the diverse nature of family routines, where participants reported \mbox{\textit{additional}} family-specific routines spanning across 11 broad categories.
Our interpretation of these insights is that, for each participant, their attitudes towards a robot’s helpfulness vary a fair amount by routines, but the variations around their respective baselines are comparable across the three questions. However, this participant-level baseline attitude towards robot helpfulness varies significantly \textit{across} the participants, and the variations are largely unexplained by whether the participants consider a routine important, or if the routine happens frequently in their families.

This illustrates the diversity of attitudes towards integrating robotic technology into families, and calls for further broadening our understanding of the ``why'' of these responses. We argue that this survey can be used as a first step to contextualize parents' attitudes towards robot integration in their homes and routines. In future work, this survey can be useful to provide a shared reference to open up further opportunities to explore family perceptions through qualitative interviews to investigate the underlying mechanisms and factors influencing the responses and attitudes. Overall, these insights emphasize the value of FRRI as a tool to quickly and systematically identify robot-facilitated routines that can be of high priority to families in general, or used to customize integration plans for specific families.

\subsection{Design Implications}

We envision the following use cases for operationalizing FRRI: 
(1) identifying design opportunities for current family routines, (2) identifying needs and preferences for robot integration into family routines and informing future robot design decisions, and (3) expanding and applying the FRRI to surrounding fields in HRI.

\subsubsection{Identifying Family Routines to Design for}
The analysis of the responses reaffirms the complexity of the challenge of integrating robots into family routines, where different routines may be present or carried out differently in different families, and each family may require or expect a different integration route. From a technical perspective, this necessitates a variable set of features and capabilities built into the platform, that are either configurable to meet each family's need, or specifically designed for pilot families based on their requirements. In addition, from an integration perspective, providing design patterns built around activities based on each family's routines are equally critical to foster positive interactions between family and robots.

\subsubsection{Planning for future Family-Robot Integration}
These implications highlight the need for a multi-pronged approach in assessing and creating technology integration plans for families. Following the literature in conducting exploratory studies~\cite{huang2008breaking}, we propose utilizing FRRI as an initial but integral component of a more comprehensive planning. For example, designers may first distribute FRRI to collect preliminary perceptions and attitudes from participants. They may then follow-up with participants to conduct semi-structured interviews to probe for the context and underlying mechanisms regarding their initial responses. The interview guide may consists of various questions regarding a set of core routines that emerged from the survey responses, but that each interview may also be tailored to each participant to include individual questions on routines that are of particular interest based on that participant's responses. In a similar vein, FRRI could also serve as a first step of a co-design session, where aside from collecting useful information on the family's routines and attitudes, researchers would also utilize it as a shared reference to base the later discussions on, facilitating rapport building as well as co-design activities.


\subsubsection{Potential to modify and expand Family-Robot Routines Inventory}

Finally, the survey's modular design may allow researchers from surrounding fields to update and modify selective components for their particular use cases. For example, one could modify the description of the specific robot, or other technology of interest, and explore family integration plans accordingly. As another example, the inventory can also be modified to reflect family routines for an eldercare facility---instead of the home---and inform future design decisions for robotic interventions at the eldercare facility.
This flexibility may allow researchers who are interested in probing family members' initial attitudes of integrating technologies into their routines, to utilize and adapt the FRRI along the ways described.

\subsection{Limitations and Future Work}
The majority of the routine items included in FRRI are adapted from existing inventories that are from several decades ago. We have modernized the wording of some of the routine items, but as suggested by the Thematic Analysis findings from the additional routines shared by the participants, it is likely that the inventory for family and child routines could benefit from a more formal and thorough update to reflect the common and important routines that fit today's families.
As noted in the previous section, assessing parents' attitudes and perceptions on integrating technology into their family routines is complex and difficult to capture in an unsupervised survey like FRRI. Instead, the survey may allow researchers to quickly estimate a parent's initial attitudes. However, to better understand a family's needs from a social robot, and to answer questions like ``\textit{Why} does a parent find it helpful (or not helpful) to integrate a robot into a specific routine,'' follow-up interviews may be needed to more accurately interpret and contextualize survey responses. 

In addition, while our sample population for the survey consists of parents, it is also important to consider perspectives from all household members, including children, when integrating robotic technologies into families. We posit that FRRI can be utilized as an important initial component of a more comprehensive plan to understand a family's need, complemented with interviews and case studies, where parents and children alike are involved, and the researchers and family members can refer back to these initial responses to guide the discussions.

Finally, while we point out in our insights that parents' perceptions of robot helpfulness varies significantly across families, explaining such diversity is beyond the scope of the current study. Family routines and rituals, along with parents' perceptions, could vary considerably depending on their cultural background, socioeconomic status, family composition, etc. Analyzing these influences could provide more in-depth insights when working with families of specific backgrounds.

\section{Conclusion}

Motivated by addressing the challenge of sustaining adoption of robots for long-term use in families, we propose that integrating social robots into family routines and rituals can serve as a possible solution. As a first step toward this goal, in this study we developed \textbf{Family-Robot Routines Inventory} and conducted an online survey with 150 parents of 5 to 16 years-old children. Through quantitative and qualitative analysis of the survey responses, we summarize four initial insights regarding parental perspectives for the integration of robots into family routines. The insights highlight that while some broad patterns exist, families are diverse both in terms of their existing routines, and in terms of their attitudes towards integrating robots into their routines. In this paper, we demonstrate how FRRI can be used to quickly collect information on the frequency and importance of families' routines, and estimate parents' initial attitudes towards integrating robots into these routines. We also note the limitations of understanding such a complex and nuanced topic through an unsupervised survey, and propose several design implications based on FRRI. For example, the survey can be utilized as an integral design resource to refer to, when planning future robot integration into home and other scenarios alike. FRRI can be operationalized in future work by accompanying interviews, case studies, and co-design studies that involve both parents and children in the process. These applications can enable more in-depth and comprehensive understanding of families' needs and preferences, their routines, and the possible integration of robotic platforms in family life.



\section*{ACKNOWLEDGMENT}
This work was supported by the National Science Foundation award \#2312354. We would like to thank Dr. Heather Kirkorian for her support in adapting and revising the FRRI survey items. Vector art used in Figure~\hyperref[fig:teaser]{1} is attributed to \texttt{pch.vector} on Freepik.



\addtolength{\textheight}{-1.5cm}   


\printbibliography

\end{document}